%% file: main.tex
\documentclass{article}
\usepackage{spconf,amsmath}
\usepackage{makecell}
\usepackage{CJKutf8}
\usepackage{booktabs}
\usepackage{amsfonts}
\usepackage{graphicx}
\usepackage[colorlinks,
            linkcolor=blue,
            anchorcolor=blue,
            citecolor=blue]{hyperref}
\usepackage[marginal]{footmisc}

\title{An end-to-end Chinese text normalization model based on Rule-guided Flat-Lattice Transformer}

\name{Wenlin Dai$^{1,\dagger}$, Changhe Song$^{1,\dagger}$\thanks{$\dagger$ Equal contributions.}, Xiang Li${^1}$, Zhiyong Wu$^{1,2,*}$\thanks{* Corresponding author.}, Huashan Pan$^3$, Xiulin Li$^3$, Helen Meng$^{1,2}$}
\address{
    $^1$ Tsinghua-CUHK Joint Research Center for Media Sciences, Technologies and Systems, \\
    Shenzhen International Graduate School, Tsinghua University, Shenzhen, China\\
    $^2$ Department of Systems Engineering and Engineering Management, \\
         The Chinese University of Hong Kong, Hong Kong SAR, China\\
    $^3$ Databaker (Beijing) Technology Co., Ltd, Beijing, China\\
    \small{
        \{dwl20, sch19, xiang-li20\}@mails.tsinghua.edu.cn, 
        \{zywu, hmmeng\}@se.cuhk.edu.hk,
        \{panhuashan, lixiulin\}@data-baker.com
    }
}

\begin{document}

\maketitle
\ninept
\input{abstract}

\input{keywords}

\input{introduction}

\input{methodology}

\input{experiment}

\input{conclusion}

\input{acknowledgement}
\bibliographystyle{IEEEbib}
\bibliography{refs}

\end{document}

%% file: abstract.tex
\begin{abstract}
Text normalization, defined as a procedure transforming non-standard words to spoken-form words, is crucial to the intelligibility of synthesized speech in text-to-speech system.
Rule-based methods without considering context can not eliminate ambiguation, 
whereas sequence-to-sequence neural network based methods suffer from the unexpected and uninterpretable errors problem.
Recently proposed hybrid system treats rule-based model and neural model as two cascaded sub-modules, where limited interaction capability makes neural network model cannot fully utilize expert knowledge contained in the rules.
Inspired by \textbf{F}lat-\textbf{LA}ttice \textbf{T}ransformer (FLAT), we propose an end-to-end Chinese text normalization model, 
which accepts Chinese characters as direct input 
and integrates expert knowledge contained in rules into the neural network, 
both contribute to the superior performance of proposed model for the text normalization task.
We also release a first publicly accessible large-scale dataset for Chinese text normalization.
Our proposed model has achieved excellent results on this dataset.

\end{abstract}

%% file: keywords.tex

%% file: introduction.tex
\section{Introduction}

Text normalization (TN) is crucial to the intelligibility of synthesized speech in text-to-speech (TTS) system.
It is defined as a procedure that transforms non-standard words (NSWs),
e.g.
written-form
numbers, symbols or characters, to spoken-form words (SFWs), 
such as 
transforming ``3.4" to ``three point four" and ``2021/10" to ``October Twenty Twenty-one".
To deal with the ambiguity problem in transforming NSWs to SFWs,
context information and NSW's inherent special construct should be considered.
For example, 
context can decide whether to read ``2021'' as year or number, 
whereas special construct of ``172.0.0.1'' can be determined as IP address.
Furthermore, to form a context, word information is 
crucial.
Take ``2021 \begin{CJK*}{UTF8}{gbsn}光年\end{CJK*}'' as example, the ``2021'' will be read as number only if the word ``\begin{CJK*}{UTF8}{gbsn}光年\end{CJK*} (light-years)'' is correctly identified;
otherwise, ``2021'' might be read as year if the keyword ``\begin{CJK*}{UTF8}{gbsn}年\end{CJK*} (year)'' is 
inaccurately matched.

Based on the taxonomy approach for NSW \cite{richard2001}, the TN tasks can be resolved by rule-based approaches which utilize handcrafted regular expressions and/or keywords \cite{2009A, 2009Document, 2016Text} to determine the category of NSWs and then convert to corresponding SFWs with predefined conversion functions.
However, the 
selection
of keywords as well as the construction of regular expression rules are time-consuming and labor-intensive.
Several machine learning methods have been proposed for the disambiguation task of NSWs, including finite state automata (FSA) \cite{2009A}, maximum entropy (ME) \cite{Jia2008TN}, conditional random fields (CRF) \cite{Liou2016}, etc.

With the development of deep learning technologies, using neural network to model contextual information has achieved impressive progress for TN task.
Sequence-to-sequence (seq2seq) models typically encode the written-form text representation into a state vector, and decode it into a sequence of spoken-form text output directly \cite{lusetti2018encoder}.
Long short-term memory (LSTM) and attention-based recurrent neural network (RNN) sequence-to-sequence models are well applied in English and Russian text normalization \cite{2016RNN, mansfield2019neural}.
Bi-directional LSTM or gated recurrent unit (GRU) are further utilized in both encoder and decoder \cite{zhang2019neural, sproat2016rnn, mansfield2019neural}.
However, directly applying sequence-to-sequence models to TN task may cause unexpected and uninterpretable errors caused by the model or data bias.

Recently, a hybrid TN system for Mandarin has been proposed, which combines a rule-based model based on pattern match and a multi-head self-attention based non-seq2seq neural network model,
to 
address
the corresponding shortcomings mentioned above \cite{zhang2020hybrid}.
According to the priority, NSWs are sent to the rule-based and neural models respectively. 
If the result of the neural model is of mismatched format, the NSW will be processed by the rule-based model again.
However, the hybrid system simply treats rule-based model and neural model as cascaded sub-modules serially, which may cause 
error accumulation.
It is easy to tell that rule-based model and neural network model can 
supplement
each other.
But limited interaction capability of these two cascaded sub-modules makes neural network model cannot fully utilize the expert knowledge included in the rules.

Inspired by the superior performance and the flexibility of the latest \textbf{F}lat-\textbf{LA}ttice \textbf{T}ransformer (FLAT) \cite{li2020flat},
we propose a FLAT based end-to-end Chinese TN model,
named \textbf{\textit{FlatTN}},
which can directly incorporate the expert knowledge in predefined rules into the network,
providing a novel way of leveraging the complementary advantages of the two models.

The advantages of using \textbf{\textit{FlatTN}} for the text normalization task falls into two aspects.
First, there is no need of the prerequisite word segmentation module.
FLAT can obtain all potential words in the sentence that match the specific lexicon, organize all characters and matched words to a lattice structure and flatten the lattice structure into spans \cite{li2020flat}, then send them into Transformer encoder.
The method of combining lexicon is fully independent of word segmentation, and more effective in using word information thanks to the freedom of choosing lexicon words in a context \cite{zhang2018chinese}.
Second, the NSW matching rules can be 
easily
incorporated into the model and the definition of rules is greatly simplified.
In the proposed model, rules are only adopted for the purpose of pattern match to derive all possible candidate NSWs in the input sentence.
There is no need for the rules to account for complex context matching for disambiguation task as in the conventional method.
We also release a large-scale dataset for the Chinese text normalization task, which will be open-sourced for public access soon.
Experimental results on this dataset demonstrate that our proposed model has achieved an excellent performance.

The contributions of our work 
are:
\begin{enumerate}
\item[1).] For the first time, we propose the use of flat-lattice transformer (FLAT) for the task of Chinese TN problem, enhancing the controllability and scalability of TN task.
\item[2).] We come up with 
a novel rule-guided FLAT model, that can directly incorporate the expert knowledge on the predefined rules into the neural network based model. The proposed model is also an end-to-end model which predicts the NSW categories directly from the raw Chinese characters with NSWs.
\item[3).] We release, as open-sourced resources, a Chinese text normalization dataset\footnote{https://github.com/thuhcsi/FlatTN}
with standard NSW taxonomies to eliminate the ambiguity in pronunciation of NSWs. It is a first publicly accessible dataset for the Chinese TN task.
\end{enumerate}

%% file: methodology.tex
\section{Methodology}
\label{sec:metho}

\begin{figure}[!t]
	\centering
	\includegraphics[width=0.80\columnwidth]{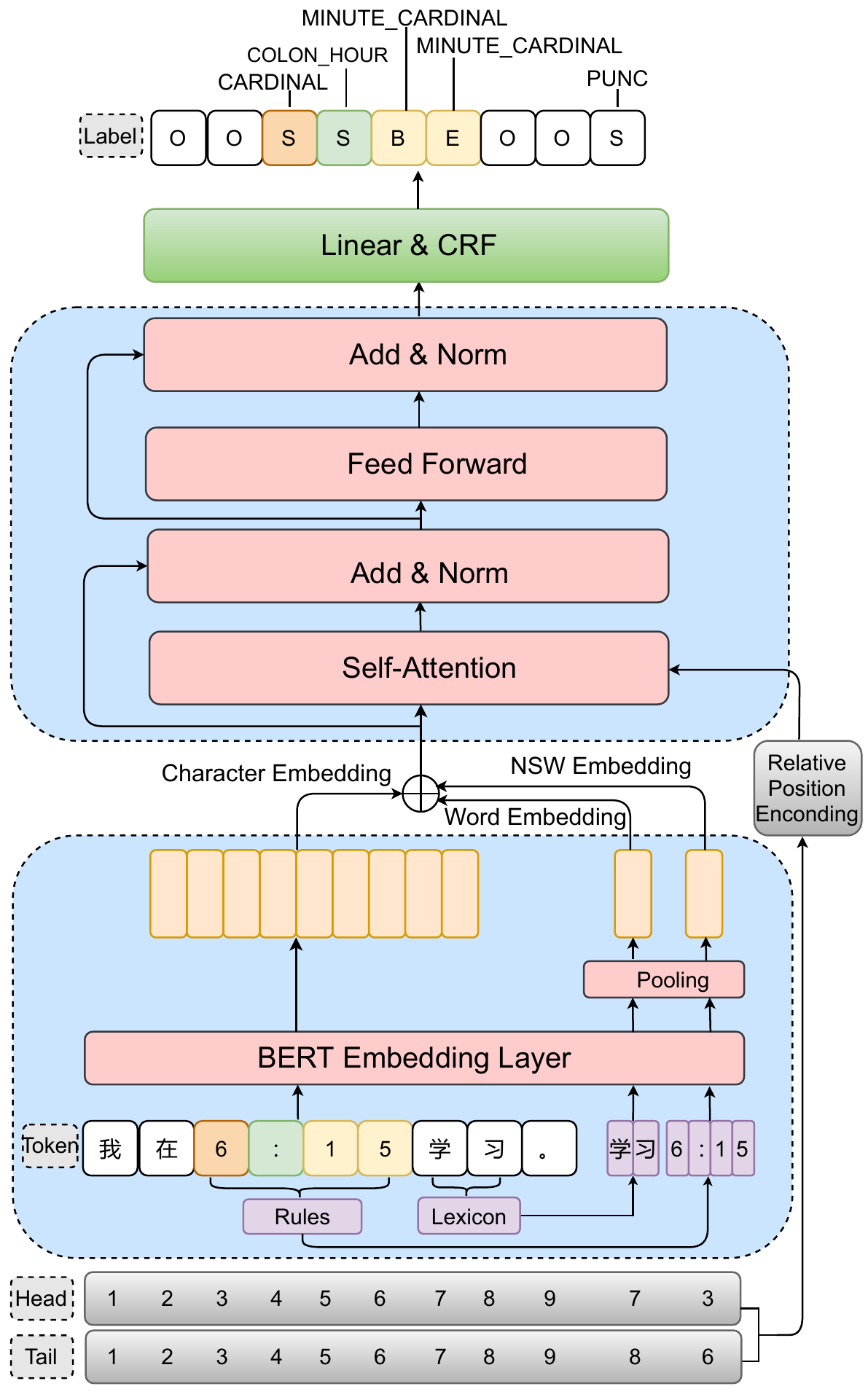}
	\caption{Proposed end-to-end model structure.}
	\label{fig:model_structure}
\end{figure}


We propose
a fully end-to-end Chinese text normalization model based on FLAT, 
which accepts characters as direct input
and can conveniently incorporate the expert knowledge from NSW matching rules.
As shown in Fig.\ref{fig:model_structure},
the model is made up of 4 parts:
(i) Lexicon and rules matching that processes the input text
and outputs a flat-lattice.
(ii) An embedding presentation layer that generates embeddings
for each token in the lattice.
(iii) A Transformer encoder that produces lattice representations
based on the generated embeddings and relative positional encodings
of all tokens.
(iv) A linear and CRF layer that predicts NSW category labels,
given the lattice representations.

For an input sentence, all the potential words that match the lexicon are obtained.
Furthermore, all the possible candidate NSWs are derived by matching the input sentence against the NSW matching rules.
The sequence of characters, the potential words and the potential NSWs in the sentence are then organized as a sequence, named flat-lattice,
where each character, word or NSW is called a token.
For each token, its head and tail positions in the original sentence are also recorded.
The character embeddings are then derived from a pretrained BERT model \cite{devlin2018bert}, from which word embeddings are obtained using a pooling layer \cite{kitaev2018constituency} for each potential word and NSW.
The character and word embeddings, together with the relative positional encoding derived from the head and tail positions for each token, are then fed into the transformer-based neural network.
The self-attention mechanism of Transformer enables embeddings to directly interact with each other \cite{vaswani2017attention}, 
from which the contextual feature representation of the lattice is obtained.
Finally, a linear layer and a conditional random field (CRF) layer are adopted to generate the categories of NSWs in the sentence.

\subsection{Lexicon and rules matching}
The model takes original characters in the sentence as its input.
The characters are then handled by two pattern matching processes.
On one hand,
a lexicon
is utilized for word matching.
On the other hand,
regular expressions, keywords, or any other form of rules
could be incorporated to search through the text for
potential NSWs.
Each matched words, NSWs, as well as the original characters
are taken as individual tokens,
and then combined into a flat-lattice.

\subsection{Embedding representation layer}
\label{ssec:metho_3_proposed}
Each token of the lattice is processed by a pre-trained BERT model \cite{devlin2018bert}
separately to obtain 
the
character embeddings.
Since BERT produces character level representations,
for words and NSWs, 
a 
pooling layer is adopted to 
obtain the final word and NSW embeddings.

\subsection{Transformer encoder with relative position encoding}
\label{ssec:metho_4_nml}
We employ a Transformer model as our encoder,
considering its strength in modeling the dependence
between arbitrary nodes.
This advantage is used to make bridges for
information exchanging among all tokens,
which is expected to improve NSWs disambiguation
by recognizing the context of individual tokens.
Moreover, the calculation process of Transformer is static,
which indicates it is agnostic to the structure of input lattice.

The Transformer encoder takes the token embeddings and relative position embeddings as its input.
The token embeddings are offered by the embedding presentation layer.
While the relative position embeddings are obtained
through a sophisticatedly designed process
proposed in FLAT \cite{li2020flat}.

\begin{figure}[h]
	\centering
	\includegraphics[width=0.7\columnwidth]{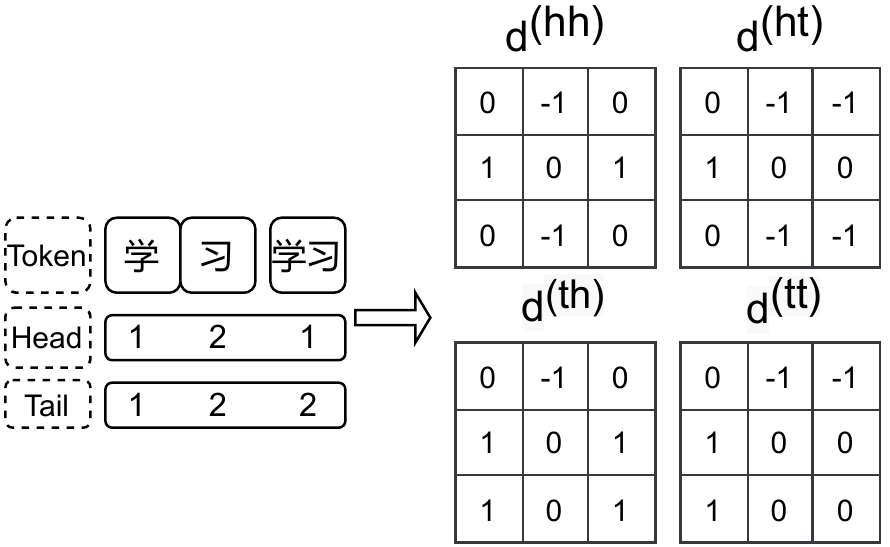}
	\caption{Relative position encoding of \begin{CJK*}{UTF8}{gbsn}``学\quad习\quad学习" \end{CJK*}.}
	\label{rpe}
\end{figure}

As shown in Fig.\ref{fig:model_structure},
the absolute position of each token in the lattice is represented
by its start and end location 
(character index) in the input sentence,
named as head and tail.
The heads and tails of all tokens
are then utilized to calculate four relative distances
between every two nodes $x_i$ and $x_j$: 
\begin{align}
\label{eq}
d_{i j}^{(h h)}&=\operatorname{head}[i]-\operatorname{head}[j]     \\
d_{i j}^{(h t)}&=\operatorname{head}[i]-\operatorname{tail}[j]     \\
d_{i j}^{(t h)}&=\operatorname{tail}[i]-\operatorname{head}[j]      \\
d_{i j}^{(t t)}&=\operatorname{tail}[i]-\operatorname{tail}[j]
\end{align}
where $d_{ij}^{(hh)}$ denotes the distance between head of $x_{i}$ and head of $x_{j}$, and other $d_{ij}^{(ht)}$, $d_{ij}^{(th)}$, $d_{ij}^{(tt)}$ have similar meanings.
Fig.\ref{rpe} shows an example of four relative position matrixes calculated with head and tail information of tokens in lattice \begin{CJK*}{UTF8}{gbsn}“学\quad习\quad学习” \end{CJK*}.

To get the final relative position encoding, a non-linear transformation is applied to the four relative distances:
\begin{equation}
\mathbf{R}_{i j}=\operatorname{ReLU}\Big \lvert(W_{r}\Big \lvert(\mathbf{p}_{d_{i j}^{(h h)}} \oplus \mathbf{p}_{d_{i j}^{(t h)}} \oplus \mathbf{p}_{d_{i j}^{(h t)}} \oplus \mathbf{p}_{d_{i j}^{(tt)}}\Big \lvert)\Big \lvert)
\end{equation}
where $W_{r}$ is a learnable parameter, $\oplus$ denotes the concatenation operator, and $\mathbf{p}_{d}$ is calculated by following equations \cite{vaswani2017attention}:
\begin{align}    
\mathbf{p}_{d}^{(2 k)} &= \sin  \left(d / 10000^{2 k / d_{\text {model }}} \right)   \\
\mathbf{p}_{d}^{(2 k+1)} &= \cos  \left(d / 10000^{2 k / d_{\text {model }}} \right)  
\end{align}
where $d$ is one of $d_{ij}^{(hh)}$, $d_{ij}^{(ht)}$, $d_{ij}^{(th)}$, $d_{ij}^{(tt)}$ and $k$ denotes the index of dimension of position encoding.

Then the relative position encoding with fused information are sent to
the Transformer encoder, together with token embeddings.
Original Transformer calculates self-attention with absolute position coding \cite{2019Character}.
In FLAT, Transformer encoder calculates self-attention with token embeddings and relative position encoding using following equation \cite{dai2019transformer}:
\begin{equation}
\begin{aligned}
\mathbf{A}_{i, j}^{*} &=\mathbf{W}_{q}^{\top} \mathbf{E}_{x_{i}}^{\top} \mathbf{E}_{x_{j}} \mathbf{W}_{k, E}+\mathbf{W}_{q}^{\top} \mathbf{E}_{x_{i}}^{\top} \mathbf{R}_{i j} \mathbf{W}_{k, R} \\
&+\mathbf{u}^{\top} \mathbf{E}_{x_{j}} \mathbf{W}_{k, E}+\mathbf{v}^{\top} \mathbf{R}_{i j} \mathbf{W}_{k, R}
\end{aligned}
\end{equation}
where $\mathbf{W}_{q}, \mathbf{W}_{k, R}, \mathbf{W}_{k, E} \in \mathbb{R}^{d_{\text {model }} \times d_{\text {head }}}$ and $\mathbf{u}, \mathbf{v} \in \mathbb{R}^{d_{\text {head }}}$ are learnable parameters,
and $\mathbf{E}_{x_{i}}, \mathbf{E}_{x_{j}} \in \mathbb{R}^{L \times d_{\text {model }}} $ are token embeddings
($d_{\text{model}}=H\times d_{\text{head}}$,
$H$ is the number of attention heads,
$d_{\text{head}}$ is the dimension of each head).

\subsection{Linear and CRF layer}
With the contextual feature representation from Transformer encoder,
a linear and CRF layer are included as decoder to
predict entity labels as the final output of our model.
CRF can obtain an optimal prediction sequence through the relationship between adjacent labels and help to reduce error occurrence.

%% file: experiment.tex

\section{Chinese TN dataset}
\begin{figure}[!ht]
	\centering
	\includegraphics[width=0.78\columnwidth]{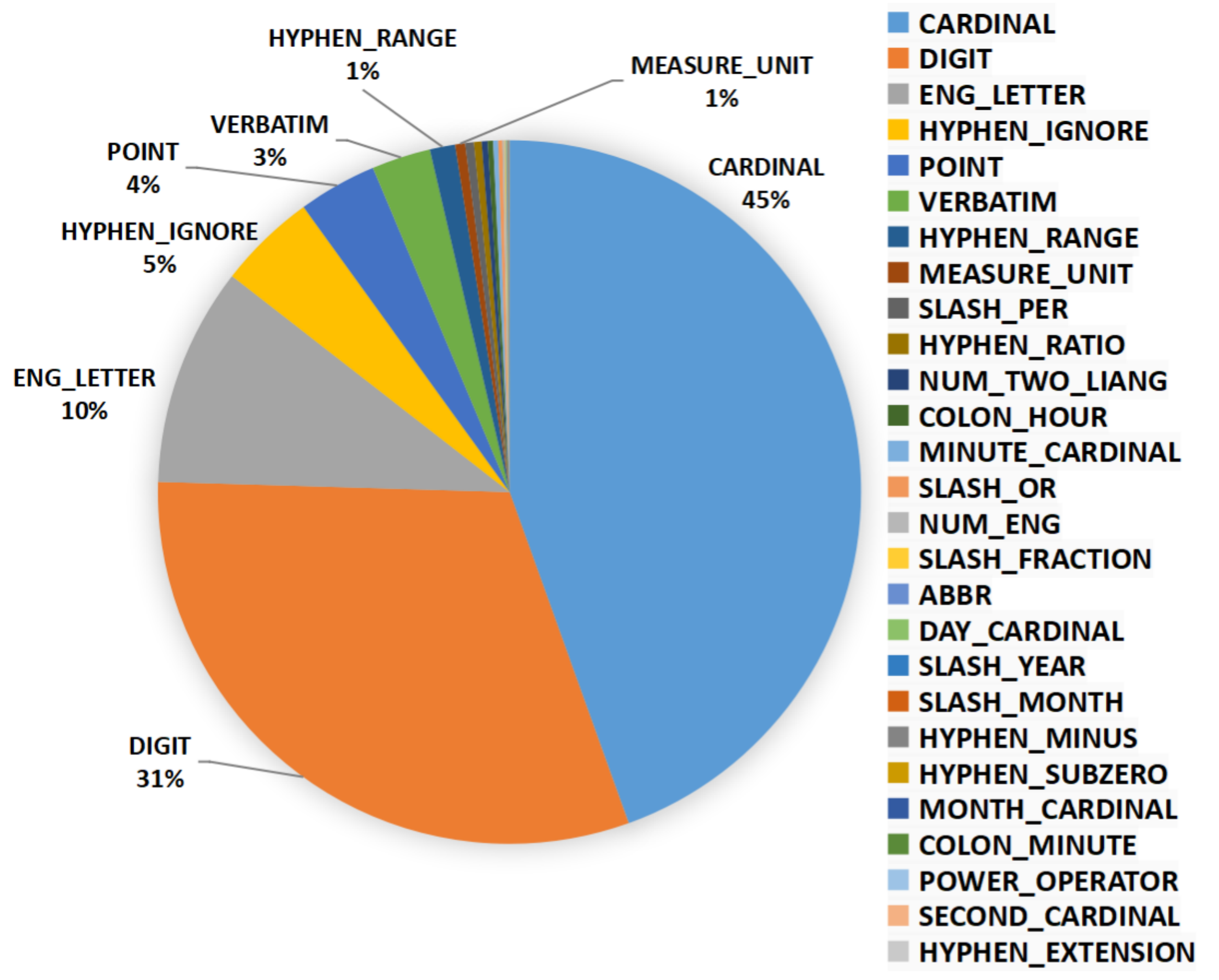}
	\caption{Proportion of NSW categories in the proposed dataset. (``O" and ``PUNC" excluded)}
	\label{fig2}
\end{figure}

In this work,
we release a large-scale Chinse TN dataset,
which is the first large-scale open source Chinse text normalization dataset
to the best of our knowledge.
A well-designed Chinese NSW classification standard is proposed along with the data,
which is made up by a total of 29 categories.
As shown in Table.\ref{t1},
each category corresponds to a handcrafted conversion function for SFW generation.
For example, a NSW likes ``2021" belongs to the ``DIGIT" category,
and should be converted into SFW as \begin{CJK*}{UTF8}{gbsn}``二零二一"\end{CJK*} (``two-zero-two-one")
by the Read$\_$DIGIT function.
Specifically,
ordinary characters that do not require transformation
are labeled as ``O" category,
which are kept intact during conversion.
And punctuation marks are labeled as ``PUNC" category,
which are simply dropped during conversion.
The original text data in the dataset are extracted from Chinese Wikipedia,
which come out as
30,000 sentences with average length of 50 characters.
Detailed NSW category labels distribution of our dataset is depict as Fig.\ref{fig2}.

The proposed Chinese TN dataset and NSW classification standard
build up a benchmark
for Chinese text normalization task.
Future researches are relieved from
the burden of NSW label and SFW converson function design,
but focus on making steady improvements based on the same foundation.

\begin{table}[!ht]
\caption{Category set of the Chinese TN dataset.}
\footnotesize
\label{t1}
\begin{tabular}{l|l|l}
\hline
Category          & How to read       & Example           \\ \hline
O              & Self-reading       &\begin{CJK*}{UTF8}{gbsn}\textbf{你好}\end{CJK*}. (\begin{CJK*}{UTF8}{gbsn}你\end{CJK*}) \\
CARDINAL       & Read cardinal       &11 pears. (11) \\
DIGIT          & One by one        & Call 911. (911)  \\
PUNC         &   No reading       & See you. (.) \\
ENG\_LETTER    & One by one        & NBA. (NBA)      \\
HYPHEN\_IGNORE & No reading       & See-you. (-)  \\
POINT          & Read as ``di\v{a}n" (\begin{CJK*}{UTF8}{gbsn}\textbf{点}\end{CJK*})     & PI is 3.14. (.)       \\
VERBATIM       & One by one        & I like C++. (++)     \\
HYPHEN\_RANGE  & Read as ``d\`{a}o" (\begin{CJK*}{UTF8}{gbsn}\textbf{到}\end{CJK*})      & July 12-20. (-)    \\
MEASURE\_UNIT  & Read the unit     & It is 24cm. (cm)    \\ 
SLASH\_PER        & Read as ``m\v{e}i" (\begin{CJK*}{UTF8}{gbsn}\textbf{每}\end{CJK*})          & 100\$/Year. (/)    \\
HYPHEN\_RATIO     & Read as ``bǐ" (\begin{CJK*}{UTF8}{gbsn}\textbf{比}\end{CJK*})           & Score is 3:2. (:)     \\
NUM\_TWO\_LIANG   & Read as ``li\v{a}ng" (\begin{CJK*}{UTF8}{gbsn}\textbf{两}\end{CJK*})        & 2\begin{CJK*}{UTF8}{gbsn}个人\end{CJK*}. (2)              \\
COLON\_HOUR       & Read as ``di\v{a}n" (\begin{CJK*}{UTF8}{gbsn}\textbf{点}\end{CJK*})         & At 9:10am. (:)        \\
MINUTE\_CARDINAL  & Add ``f\={e}n" (\begin{CJK*}{UTF8}{gbsn}\textbf{分}\end{CJK*})             & At 9:10am. (10)       \\
SLASH\_OR         & Read as ``hu\`{o}" (\begin{CJK*}{UTF8}{gbsn}\textbf{或}\end{CJK*})      & Apple/pear. (/)                             \\
NUM\_ENG          & Read as English   & Seq2seq. (2)                                \\
SLASH\_FRACTION   & Read as fraction  & 3/4. (/)                           \\ 
ABBR              & Abbreviation      & Mr Smith. (Mr)                              \\
DAY\_CARDINAL     & Add ``r\`{i}" (\begin{CJK*}{UTF8}{gbsn}\textbf{日}\end{CJK*})          & 2021/09/06. (06)                            \\
SLASH\_YEAR       & Read as ``ni\'{a}n" (\begin{CJK*}{UTF8}{gbsn}\textbf{年}\end{CJK*})    & 2021/09. (/)                               \\
SLASH\_MONTH      & Add ``yu\`{e}" (\begin{CJK*}{UTF8}{gbsn}\textbf{月}\end{CJK*})         & 09/06. (/)                                  \\
HYPHEN\_MINUS     & Read as ``f\`{u}" (\begin{CJK*}{UTF8}{gbsn}\textbf{负}\end{CJK*})      & -5. (-)                                    \\
HYPHEN\_SUBZERO   & Read ``l\'{i}ng xi\`{a}" (\begin{CJK*}{UTF8}{gbsn}\textbf{零下}\end{CJK*})   & -20℃. (-)                                   \\
MONTH\_CARDINAL   & Add ``yu\`{e}" (\begin{CJK*}{UTF8}{gbsn}\textbf{月}\end{CJK*})         & 2021/09. (09)                              \\
COLON\_MINUTE     & Read as ``f\={e}n" (\begin{CJK*}{UTF8}{gbsn}\textbf{分}\end{CJK*})     & 59:20. (:)                                  \\
SECOND\_CARDINAL  & Add ``mi\v{a}o" (\begin{CJK*}{UTF8}{gbsn}\textbf{秒}\end{CJK*})        & 23:59:20. (20)                              \\
HYPHEN\_EXTENSION & Read as ``zhu\v{a}n" (\begin{CJK*}{UTF8}{gbsn}\textbf{转}\end{CJK*})   & 12345-678. (-)                              \\
POWER\_OPERATOR   & Read ``c\`{i} f\={a}ng" (\begin{CJK*}{UTF8}{gbsn}\textbf{次方}\end{CJK*})    & 2\textasciicircum{}3. (\textasciicircum{}) \\ \hline

\end{tabular}
\end{table}

\section{Experiment}
\label{sec:experiment}

\subsection{Experiment setup}

We split all sentences into characters and label them with ``BMESO" format, and then randomly separate the dataset into training set, verfication set, and test set at a ratio of 8:1:1.
To compare our proposed model with rule based model and mainstream models based on neural network, three baseline models are constructed for comparison:
\begin{enumerate}
\item[1).] \textbf{Rule-based}: The adopted rule-based system is a commercial system by Databaker. 
Regular expressions are firstly used to match the NSW candidates, and then predefined rules are used to disambiguate the category for each NSW. 
\item[2).] \textbf{BERT-MLP}: A neural network that obtains character embeddings using BERT, and predicts NSW labels using multi-layer perception \cite{2019Research} followed by CRF.
\item[3).] \textbf{BERT-LSTM}: A neural network similar to BERT-MLP, but with perception layer replaced by LTSM layer \cite{2019Research}.
\end{enumerate}

\subsection{Model performance}

\subsubsection{Overall performance}

Experiment results confirm that the proposed FlatTN model reaches
the best results on both accuracy and F1-score.
As revealed in Table.\ref{t2}, BERT-LSTM outperforms all other baseline models, due to its sequential structure.
And there exists a noticeable gap
between the Rule-based model and other models,
which indicates the significance of pre-trained language models and neural modules
in the promotion of text normalization performance.

\begin{table}[!ht]
\centering
\footnotesize
\caption{Accuracy and F1 of different models.}
\label{t2}
\begin{tabular}{|c|c|c|}

\hline
Method & Accuracy &F1  \\ \hline
Rule-based  &  0.8775  &   0.8729   \\
BERT-MLP  &  0.9869   & 0.9580          \\
BERT-LSTM  &  0.9885   &  0.9638     \\
\textbf{FlatTN} & \textbf{0.9907}  &  \textbf{0.9708}      \\ \hline

\end{tabular}
\end{table}

\subsubsection{Model performance regarding different categories} 
To explore the classification performance on different categories, our model is evaluated further in terms of precision, recall and F1-score on all categories.
The results of 10 typical categories are shown in Table \ref{t3}. 
It is clear that the categories frequently appear in Chinese text are more likely to obtain good performance, while the categories rarely appear in Chinese text may get worse performance than others.
\begin{table}[!ht]
\footnotesize
\centering
\caption{Model performance on the test set.}
\label{t3}
\begin{tabular}{|c|c|c|c|}
\hline
Category              & Precision & Recall  &\quad F1        \\ \hline
PUNC               &       0.9952 &  0.9978    &      0.9965       \\
MINUTE\_CARDINAL             &   0.9706     &     1.0000  &    0.9851     \\
POINT           &    0.9610       &    0.9769    &      0.9689    \\
CARDINAL        &    0.9676      &  0.9607    &       0.9641     \\
DIGIT                 &   0.9487     &     0.9567  &    0.9527    \\
SLASH\_PER           &   0.8889     &     1.0000     &    0.9412  \\
HYPHEN\_RATIO       &   0.9677     &     0.9091     &  0.9375 \\
VERBATIM              &     0.9385   &   0.8750      &   0.9057 \\
HYPHEN\_RANGE           &    0.7876    &    0.9468     &   0.8599  \\ 
HYPHEN\_IGNORE   	&         0.8386	 & 0.8469	&  0.8428\\

\hline
\end{tabular}
\end{table}

\subsubsection{Ablation study}

Ablation study of the proposed FlatTN model is established by
removing lexicon, or rules, or both of them from
the input process of the proposed model.
As shown in Table.\ref{t4}, both lexicon and rules can help to
improve results on accuracy and F1-score ,
and the performance promotion of the each part
is complementary to the other.
This is consistent with our expectation that rule-guided FlatTN
should present better text normalization result,
since it is embodied with both context information modeling and expert knowledge.

\begin{table}[!ht]
\centering

\footnotesize
\caption{Results of ablation experiment.}
\label{t4}
\begin{tabular}{|c|c|c|}

\hline
Method & Accuracy& F1 \\ \hline
         
    FlatTN  &    \textbf{0.9907} & \textbf{0.9708}     \\
    \hspace{2mm}- Lexicon&  0.9886 & 0.9658    \\
    \hspace{2mm}- Rules &   0.9880 & 0.9632   \\
    \hspace{4mm}- Lexicon & 0.9879 & 0.9596 \\ \hline
\end{tabular}
\end{table}


%% file: conclusion.tex
\section{Conclusion}
\label{sec:conclusion}
In this paper, we propose an end-to-end Chinese text normalization model based on rule-guided flat-lattice Transformer.
Our model combines the scalability and flexibility of rules and the ability to model context and utilize data efficiently of Transformer encoder.
We also release a first publicly accessible large-scale dataset for Chinese text normalization task.
Our proposed model achieves the accuracy of 99.1$\%$ on NSW classification on the test set, achieving better performance compared with the methods using rules or neural networks alone, and the ablation experiment proves the importance of rules and lexicon in our model.

%% file: acknowledgement.tex

\textbf{Acknowledgement}: This work is partially supported by 
National Natural Science Foundation of China (NSFC) (62076144),
the Major Project of National Social Science Foundation of China (NSSF) (13\&ZD189).
